\documentclass[letterpaper, 10 pt, conference]{ieeeconf}  

\IEEEoverridecommandlockouts                              

\usepackage{graphicx} 
\usepackage{epsfig} 
\usepackage{times} 
\usepackage{amsmath} 
\usepackage{amssymb}  
\usepackage{bm}
\usepackage{subfig}
\usepackage{url}
\usepackage[monochrome]{color}

\DeclareMathOperator*{\argmin}{argmin}
\title{\LARGE \bf
	Efficient two step optimization for large embedded deformation graph based SLAM
}

\author{Jingwei Song$^{1}$, Fang Bai$^{1}$, Liang Zhao$^{1}$, Shoudong Huang$^{1}$ and Rong Xiong$^{2}$}

\begin{document}

	\maketitle
	\thispagestyle{empty}
	\pagestyle{empty}

	\begin{abstract}
		Embedded deformation graph is a widely used technique in deformable geometry and graphical problems. Although the technique has been transmitted to stereo (or RGBD) sensor based SLAM applications, it remains challenging to compromise the computational cost as the model grows. In practice, the processing time grows rapidly in accordance with the expansion of maps. In this paper, we propose an approach to decouple nodes of deformation graph in large scale dense deformable SLAM and keep the estimation time to be constant. We observe that only partial deformable nodes in the graph are connected to visible points. Based on this fact, the sparsity of the original Hessian matrix is utilized to split parameter estimation into two independent steps. With this new technique, we achieve faster parameter estimation with amortized computation complexity reduced from $O(n^2)$ to closing $O(1)$. As a result, the computation cost barely increases as the map keeps growing. Based on our strategy, computational bottleneck in large scale embedded deformation graph based applications will be greatly mitigated. The effectiveness is validated by experiments, featuring large scale deformation scenarios. 	
	\end{abstract}

	\section{INTRODUCTION}
	Simultaneous Localization and Mapping (SLAM) is a technique widely used in robotics for pose estimation and environment mapping. While SLAM  in the rigid scenario is quite mature, SLAM with deformations is still new to the community. Several attempts in this direction include RGBD camera based deformable human body reconstruction and RGB camera based soft-tissue SLAM in Minimally Invasive Surgery (MIS). The key challenge \textcolor{red}{in} developing deformable SLAM is how to define and formulate the deformation. \par 
	
	\textcolor{red}{As an early stage, researchers assume deformation scenario is technically similar to SLAM in dynamic environment. By assuming a large amount of target features to be static, approaches like \cite{grasa2011ekf} and \cite{lin2013simultaneous} were proposed, adopted conventional extended Kalman filter (EKF) SLAM and Parallel Tracking and Mapping (PTAM) by applying} some thresholds \textcolor{red}{to separate} rigid and non-rigid feature points. More \textcolor{red}{recent work ORB-SLAM} \cite{mur2015orb} have been \textcolor{red}{proposed} and modified in \cite{mahmoud2016orbslam} \cite{mahmoud2017slam}  \cite{chen2018slam} \cite{marmol2019dense}. \textcolor{red}{The above works show that extending traditional rigid SLAM to non-rigid scenarios is possible.} \textcolor{red}{However,} map reconstruction based on the rigid assumption inevitably leads to gaps within rigid patches from different \textcolor{red}{perspectives}. The more extensive deformation of the environment, the larger gaps exist in the recovered map.\par 
	
	\textcolor{red}{A better solution is to model the deformation.} Recognizing map deformation modeling is vital \textcolor{red}{and indeed many} researches turn to \textcolor{red}{the} computer vision society for efficient deformation description. \textcolor{red}{One of the most well-known and well-studied techniques is the Embedded Deformation (ED) \cite{sumner2007embedded}, which was initially proposed for designing smooth character motions in cartoons.} ED graph makes use of a sparsely interconnected scattered nodes with attributes of local rigid transformation. By mixing these local rigid transformations with weights, a global deformation \textcolor{red}{is} simulated in a discrete but smooth form. ED graph has been widely applied in RGBD \textcolor{red}{and} stereo SLAM \cite{whelan2016elasticfusion} \cite{song2018dynamic} and 3D human reconstructions \cite{newcombe2015dynamicfusion} \cite{innmann2016volumedeform} \textcolor{red}{(discrete hierarchical warp grid but very similar to ED graph)} \cite{dou2016fusion4d} \cite{guo2017real} \cite{dou2017motion2fusion}. \textcolor{red}{The ED graph based} SLAM formulation is able to describe \textcolor{red}{the deformation of the environment} in addition to camera pose, \textcolor{red}{which blossoms with the application} like \cite{newcombe2015dynamicfusion} \cite{innmann2016volumedeform} \cite{dou2016fusion4d} \cite{dou2017motion2fusion} \textcolor{red}{for} human motion reconstruction \textcolor{red}{and} \cite{song2018dynamic} \textcolor{red}{for} large scale SLAM in MIS.\par 
	
	\textcolor{red}{In the meantime, for more challenging monocular equipments, Finite Element Method (FEM) \cite{agudo20123d} \cite{petit2018environment} and Structure from Template (SfT) \cite{Lamarca_2018_ECCV_Workshops} \cite{lamarca2019defslam} were proposed to simulate deformation. Both methods use a predefined grid or triangular mesh template to describe the soft-tissue. However, to the best of our knowledge, it is hard to apply these methods when the map is incrementally built and no complete real-time implementation has demonstrated how it will effectively be applied in large scenarios.} \par 
	
	
	
	Overall, ED graph based formulation is the most applicable and widely used approach for modeling deformations. Numerous real-time dense SLAM systems have validated its effectiveness\textcolor{red}{, particularly in batch scenarios, with} fast sequential or parallel implementation.    
	\textcolor{red}{However, ED graph based formulation also comes with disadvantages and limitations.    One major issue in ED graph is that when a new observation is incorporated, the number of nodes increases dramatically, posing heavy computational burden. Little attention has been paid in the field of truncated signed distance function based 3D human reconstruction because the target size, as well as map extent, are predefined. State estimation and map updating are all confined within a volume. In more general cases, however, as reported in \cite{song2018dynamic}, when reconstructing geometry without a predefined volume, the environment size is unbounded due to the sharp growth of the graph, which eventually implies an amortized $O(n^2)$ complexity with respect to the number of nodes in the graph. In a word, optimizing an expanding ED graph in an unconstrained space significantly limits the performance of the system.} \par 
	
	\textcolor{red}{Even though little is known in speeding up ED graph based SLAM systems, numerous publications are dedicated to the graph based optimization in rigid scenarios. A typical graph optimization problem models robot poses (or landmark positions) as nodes in the graph, while the edges encode the relative measurement between connected nodes \cite{thrun2005probabilistic}. Graph sparsification is the most widely applied technique to marginalize subsets of nodes \cite{carlevaris2014generic} \cite{eckenhoff2016decoupled} \cite{vallve2018graph}. The key process in this topic is to sparsify edges and marginalize nodes based on indicators like divergence \cite{mackay2003information}. We will show in this paper that marginalization methods are of great value to enhance efficiency in ED based SLAM. In this paper, we will mitigate the $O(n^2)$ computational bottleneck encountered in the expanding environment, for ED graph based SLAM as reported in \cite{song2018dynamic}. We analyze the spatial relationship between ED graph and observation and reveal the inherent sparsity pattern of the Hessian matrix. With this discovery, we classify ED nodes into points relevant (\textbf{PR}) and points irrelevant (\textbf{PI}) nodes and propose a decoupled optimization strategy.} \par 
	
	\textcolor{red}{In this work we convert the ED graph based formulation into a matrix form, which unveils a sparsity pattern that can be further exploited by restructuring the Jacobian matrix to benefit efficient marginalization.}
	Based on \textcolor{red}{this insight}, we \textcolor{red}{develop} a method which split ED graph optimization into two steps and a lossy decoupled optimization strategy is proposed. The accuracy and effectiveness are tested \textcolor{red}{on both the} ED based geometry deformation as well as ED based SLAM application. \textcolor{red}{The proposed strategy is validated on the MIS-SLAM framework} \cite{song2018dynamic} with and without the proposed strategy. \textcolor{red}{For more details on the MIS-SLAM framework, please refer to the flowchart of in \cite{song2018dynamic}.}

	\begin{figure}
		\centering 
		\includegraphics[width=0.4\textwidth]{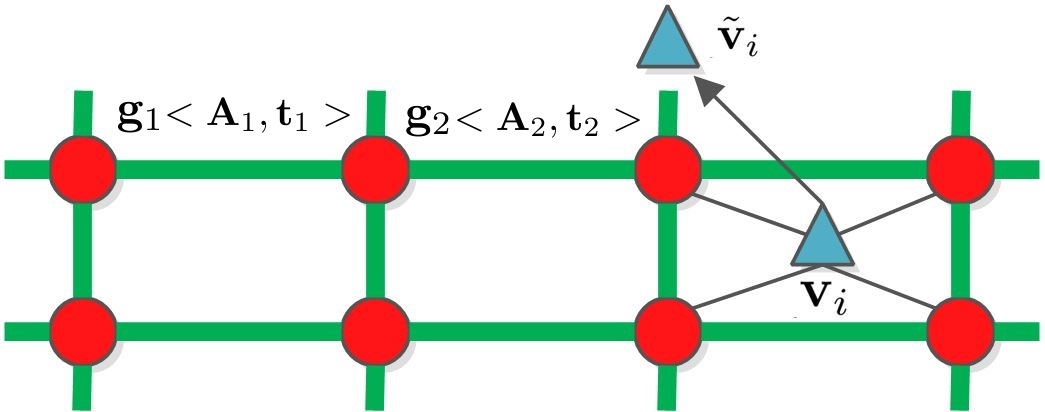}
		\caption{\textcolor{red}{
		An example of ED graph. The red circles are ED nodes, say node $j$, encoding a geometric position $\mathbf{g}_j$, and an affine transformation given by $\mathbf{A}_j$ and $\mathbf{t}_j$.
	The blue triangle is a vertex, that can be deformed from $\mathbf{v}_i$ to $\tilde{\mathbf{v}}_i$, through the impact of its neighboring ED nodes.} }
		\label{fig:deformation_graph}
	\end{figure}
	\section{Revisit general ED graph based SLAM}
	A ED graph, as other graph representations, comprises nodes and edges. See Fig. \ref{fig:deformation_graph} for an example. We use the term `node' to describe a ED node, and `vertex' to describe a feature point. Each ED node $j$ defines a local rigid motion modeled by ${\mathbf{A}_j, \mathbf{t}_j}$, \textcolor{red}{where} $\mathbf{A}_j$ $\in\mathbb{R}^{3\times3}$ is an affine matrix and $\mathbf{t}_j$ $\in\mathbb{R}^3$ is a translation vector. Given a vertex $\mathbf{v}_i$, neighbouring ED nodes define a deformation field indicating how $\mathbf{v}_i$ is deformed to target position $\mathbf{\tilde{v}}_i$. For a mesh based geometry, \textcolor{red}{shape deformation is equivalent to translate vertices in the ED graph}.\par 

	
	In practice, to avoid over-parameterization, the ED node $\mathbf{g}_j$ are deliberately positioned to minimize the size of parameters. For example in \cite{sumner2007embedded} \cite{newcombe2015dynamicfusion} \cite{dou2016fusion4d} \cite{song2018mis}, ED nodes are generated by uniformly downsampling the original \textcolor{red}{model}. \textcolor{red}{The deformation between the point $\mathbf{v}_i$ and $\tilde{\mathbf{v}}_i$ is modeled as a transformation in the following form:}\par
	\begin{equation}
	\tilde{\mathbf{v}}_i=\mathbf{R}_c{\sum_{j=1}^m \omega_j(\mathbf{v}_i)[\mathbf{A}_j(\mathbf{v}_i-\mathbf{g}_j)+\mathbf{g}_j+\mathbf{t}_j]}+\mathbf{T}_c,
	\label{TransformationFomulation}
	\end{equation}
	where $\mathbf{R}_c$ and $\mathbf{T}_c$ are global rotation and translation relating to camera \textcolor{red}{motion}, $m$ is \textcolor{red}{the} number of neighboring nodes. The global camera pose $[\mathbf{R}_c, \mathbf{T}_c]$ is not considered in the original ED work \cite{sumner2007embedded}, however added later on in \cite{newcombe2015dynamicfusion} \cite{dou2016fusion4d} \cite{song2018mis}. 
	$\omega_j(\mathbf{v}_i)$ is \textcolor{red}{a weight that quantifies the contribution of the ED node $j$ to the vertex $\mathbf{v}_i$.} We choose the number of neighboring nodes to $m=4$ and define the weight:
	\begin{equation}
	\label{eq_weight}
	\omega_j(\mathbf{v}_i)=1-||\mathbf{v}_i-\mathbf{g}_j||/d_{max},
	\end{equation}
	where $d_{max}$ is the maximum distance \textcolor{red}{from the vertex $\mathbf{v}_i$ to the neighbouring} $k + 1$ nearest ED node.

	Eq. (\ref{TransformationFomulation}) shows how \textcolor{red}{one vertex} is transformed with \textcolor{red}{the ED graph}. Conversely, if the source ($\mathbf{v}_i$) and target ($\tilde{\mathbf{v}}_i$) point pairs are given, \textcolor{red}{the parameters} of ED graph ($\mathbf{A}_j$ and $\mathbf{t}_j$) can also be inferred. Thus, the ED graph ED based SLAM formulation is to infer the ED graph parameter from \textcolor{red}{a cluster of} arbitrary source and target points pairs. Then, ED graph is applied to deform the whole model. To the best of our knowledge, all ED graph based SLAM \textcolor{red}{formulation consist} of at least three terms: Rotation constraint, regularization constraint \textcolor{red}{and a penalty term on the} distance of deformed source and target point pairs. \textcolor{red}{Some methods} introduce more innovating terms, like visual hull term in \cite{dou2016fusion4d}, key features in \cite{innmann2016volumedeform} \textcolor{red}{which are beyond the scope. A ED graph based SLAM is to minimize:}\par
	\begin{equation}
	\begin{split}
	\argmin\limits_{\mathbf{R}_c,\mathbf{T}_c,\mathbf{A}_1,\mathbf{t}_1...\mathbf{A}_m,\mathbf{t}_m} &\omega_{rot}E_{rot}+\omega_{reg} E_{reg}+\omega_{data} E_{data}.
	\end{split}
	\label{energyfunction}
	\end{equation}
	\textcolor{red}{The major key point pairs matching term is $E_{data}$, while} two soft constraints $E_{rot}$ and $E_{reg}$ regulate deformation. $E_{rot}$ \textcolor{red}{hauls} the affine matrix close to $SO(3)$ by minimizing the column vectors $\mathbf{c}_1$, $\mathbf{c}_2$ and $\mathbf{c}_3$ in:
	\begin{equation}
	\label{Eq_E_rot}
	E_{rot}=\sum_{j=1}^m Rot(\mathbf{A}_j),
	\end{equation}
	\begin{equation}
	\begin{aligned}
	Rot(\mathbf{A})=(\mathbf{c}_1^T\cdot\mathbf{c}_2)^2+(\mathbf{c}_1^T\cdot\mathbf{c}_3)^2+(\mathbf{c}_2^T\cdot\mathbf{c}_3)^2+\\
	(\mathbf{c}_1^T\cdot\mathbf{c}_1-1)^2+(\mathbf{c}_2^T\cdot\mathbf{c}_2-1)^2+(\mathbf{c}_3^T\cdot\mathbf{c}_3-1)^2.
	\end{aligned}
	\end{equation}

	Regularization is to ensure smoothness of the deformed surface, the motion field generated from ED nodes \textcolor{red}{should be} consistent; otherwise, the deformed shape will be fragmented and distorted. This is a general practice named `as-rigid-as-possible' and defined in following form:\par
	\begin{equation}
	\setlength{\abovedisplayskip}{3pt}
	\setlength{\belowdisplayskip}{3pt}
	E_{reg}=\sum_{j=1}^m\sum_{k\in{\mathbb{N}(j)}} \alpha_{jk}||\mathbf{A}_j(\mathbf{g}_k-\mathbf{g}_j)+\mathbf{g}_j+\mathbf{t}_j-(\mathbf{g}_k+\mathbf{t}_k)||^2.
	\label{RegulationConstraint}
	\end{equation}
	Similar to \cite{sumner2007embedded}, $\alpha_{jk}$ is the weight quantifying the influences of the neighboring ED nodes. $\mathbb{N}(j)$ is \textcolor{red}{the set of} all connected nodes to node $j$. Normally, the number of neighboring nodes is 4.\par

	\textcolor{red}{To solve geometrical model to frame registration, `back-projection' formulation is adopted as a substitution to iterative closest point (ICP). Readers may refer to \cite{dou2016fusion4d} \cite{newcombe2015dynamicfusion} \cite{song2018mis}.} For simplicity, we introduce the basic source and target key point \textcolor{red}{pairs} described by \cite{sumner2007embedded}. Let $\mathbf{\tilde{v}}_i$ and $\mathbf{\hat{v}}_i$ be the pairs of source-target points (defined in Eq. (\ref{TransformationFomulation})). Normally predefined in interactive \textcolor{red}{phase}, these key points define how model is to be deformed. These key points define a data term which minimizes a distance as:
	
	\begin{equation}
	\begin{aligned}
	E_{data}=&\sum_{i=1}^n||\mathbf{\tilde{v}}_i-\mathbf{\hat{v}}_i||^2.
	\end{aligned}
	\label{E_dataterm}
	\end{equation}

	\vspace{0pt} 
	\hspace{0pt}
	\begin{figure*}[h]	
		\centering
		\subfloat[]{	
			\begin{minipage}[htpb]{0.18\textwidth}	
				\centering
				\includegraphics[width=1\linewidth]{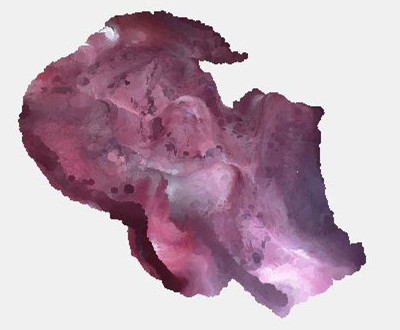}			
			\end{minipage}				
		}
		\subfloat[]{	
			\begin{minipage}[htpb]{0.18\textwidth}	
				\centering
				\includegraphics[width=1\linewidth]{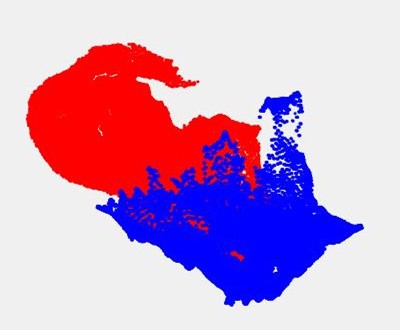}			
			\end{minipage}				
		}
		\subfloat[]{	
			\begin{minipage}[htpb]{0.18\textwidth}	
				\centering
				\includegraphics[width=1\linewidth]{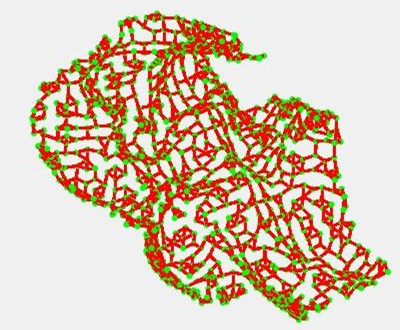}			
			\end{minipage}				
		}
		\subfloat[]{	
			\begin{minipage}[htpb]{0.18\textwidth}	
				\centering
				\includegraphics[width=1\linewidth]{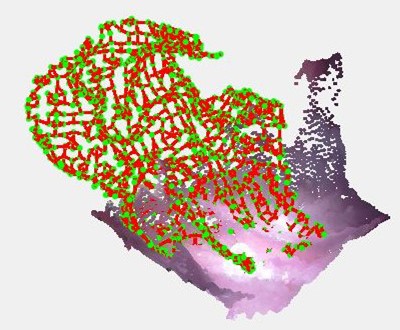}			
			\end{minipage}				
		}\/
		\caption{Illustrated are the visible points and the node graph. First is the latest reconstruction. Second shows both the model (red) and the target depth (blue). Third is the ED nodes and the edges. Last presents the ED nodes and the target depth.}
		\label{fig:node_points_relation}
	\end{figure*}

	\setlength{\abovedisplayskip}{2pt}
	\section{Efficient two step optimization}
	\subsection{Matrix \textcolor{red}{form} of ED graph deformation}
	To fully exploit the structure we rewrite the data term Eq. (\ref{E_dataterm}) \textcolor{red}{in a} matrix form for the convenience of sparsity analysis. Let us consider a group of \textcolor{red}{predefined} key source points  ${\mathbf{P}}=[\mathbf{v}_1...\mathbf{v}_n]$ and the key target points ${\mathbf{\tilde{P}}}=[\tilde{\mathbf{v}}_1...\tilde{\mathbf{v}}_n]$ \textcolor{red}{in the matrix form}. \textcolor{red}{According to Eq. (\ref{TransformationFomulation}),} each point ${\mathbf{v}_i}$ is deformed by its 4 neighboring nodes. Thus we define two matrices $\mathbf{M}$ and $\mathbf{C}$ (Eq. (\ref{MatrixMC})):
	\begin{equation}
	\label{MatrixMC}
	\begin{split}
	\underset{3m\times n}{\mathbf{M}} =
	\begin{bmatrix}
	\cdots&&\cdots\\
	\cdots&\mathbf{e}_{n,1}&\cdots\\
	\cdots&&\cdots\\
	\cdots&\mathbf{e}_{n,2}&\cdots\\
	\cdots&&\cdots\\
	\cdots&\mathbf{e}_{n,3}&\cdots\\
	\cdots&&\cdots\\
	\cdots&\mathbf{e}_{n,4}&\cdots\\
	\cdots&&\cdots\\
	\end{bmatrix}
	\end{split}
	\underset{m\times n}{\mathbf{C}} =
	\begin{bmatrix}
	\cdots&&\cdots\\
	\cdots&\omega_{\mathbf{N}(n,1)}&\cdots\\
	\cdots&&\cdots\\
	\cdots&\omega_{\mathbf{N}(n,2)}&\cdots\\
	\cdots&&\cdots\\
	\cdots&\omega_{\mathbf{N}(n,3)}&\cdots\\
	\cdots&&\cdots\\
	\cdots&\omega_{\mathbf{N}(n,4)}&\cdots\\
	\cdots&&\cdots\\
	\end{bmatrix},
	\end{equation}

	\noindent where $\mathbf{e}_{n,k}=\omega_{\mathbf{N}(n,k)}*(\mathbf{v}_i-\mathbf{g}_{\mathbf{N}(n,k)})$ and $\mathbf{N}(i,j)(j=1,2,3,4)$ is the set of neighboring node to point $\mathbf{v}_i$. In \textcolor{red}{matrices} $\mathbf{M}$ and $\mathbf{C}$, note that non-zero elements are not aligned. Each column only has 4 \textcolor{red}{non-zero} elements (neighboring nodes). \textcolor{red}{The} sum of each column in matrix $\mathbf{C}$ is 1 \textcolor{red}{due to the point to node topology.} In Eq. (\ref{E_dataterm}), a source point $\mathbf{v}_i$ is transformed by its 4 neighboring node which yields 4 non-zero elements in $\mathbf{M}$ and $\mathbf{C}$. The sum of all weight $\omega_j(\mathbf{v}_i)$ is 1. \textcolor{red}{Note} that different source points have different topology, thus the location of non-zero elements \textcolor{red}{are not aligned well in each column}. We arrange the parameters of ED nodes $\mathbf{A}_i$ and $\mathbf{t}_i$ in the following form:
	\begin{equation}
	\label{MatrixA}
	{\bm{\Lambda}} =
	\left(
	\begin{array}{ccc}
	{\mathbf{A}_1}&\cdots&{\mathbf{A}_m}.\\
	\end{array}
	\right)
	\end{equation}
	\begin{equation}
	\label{MatrixT}
	{\mathbf{T}} =
	\left(
	\begin{array}{ccc}
	{\mathbf{t}_1}+{\mathbf{g}_1}&\cdots&{\mathbf{t}_m}+{\mathbf{g}_m}\\
	\end{array}
	\right).
	\end{equation}
	
	\textcolor{red}{Then Eq. (\ref{E_dataterm}) takes the following form:}\par 
	\begin{equation}
	\label{TransformationMatrix}
	E_{data}=||{\mathrm{\mathbf{R}_c}}\cdot [{\mathrm{\mathbf{\Lambda}}}\cdot{\mathbf{M}}+{\mathbf{T}} \cdot \mathbf{C}]+{\mathrm{{\mathbf{T}_c}}} \otimes \mathbf{1}-\mathbf{\tilde{P}}||_F^2. 
	\end{equation}
	where $\otimes$ is the kronecker product. $\mathbf{1}$ is $1 \times n$ vector of ones.	\textcolor{red}{And $||\cdot||_F^2$ is the Frobenius norm. Eq. (\ref{TransformationMatrix}) can be written compactly in the following form:}
	\begin{equation}
	\label{TransformationMatrixForm}
	E_{data}=||{\mathrm{\mathbf{R}_c}}\cdot
	\left(
	\begin{array}{cc}
	{\mathrm{\mathbf{\Lambda}}}&{\mathbf{T}}\\
	\end{array}
	\right)
	\cdot
	\left(
	\begin{array}{c}
{\mathbf{M}}\\
	\mathbf{C}
	\end{array}
	\right)+{\mathrm{{\mathbf{T}_c}}} \otimes \mathbf{1}-\mathbf{\tilde{P}}||_F^2. 
	\end{equation}
	If we further define ${\bm{\Pi}}=[{\mathbf{M}}^T\  {\mathbf{C}}^T]$ and ${\mathbf{\bm{\Phi}}}=[{\mathbf{\bm{\Lambda}}}\  {\mathbf{T}}]^T$, Eq. (\ref{TransformationMatrixForm}) can be formed as follows:
	\begin{equation}
	\label{TransformationMatrixFinalForm}
	E_{data}=||{\mathrm{\mathbf{R}_c}}[{\bm{\Pi}} {\mathbf{\bm{\Phi}}}]^T+{\mathrm{{\mathbf{T}_c}}} \otimes \mathbf{1} -\mathbf{\tilde{P}}||_F^2.
	\end{equation}

The property of Jacobian of $E_{data}$ is \textcolor{red}{determined by ${\bm{\Pi}}$.} 
	
	\subsection{Sparsity in ED graph formulation}
	\label{session:Sparsity in ED node}

\begin{figure}
		\centering 
		\includegraphics[width=0.44\textwidth]{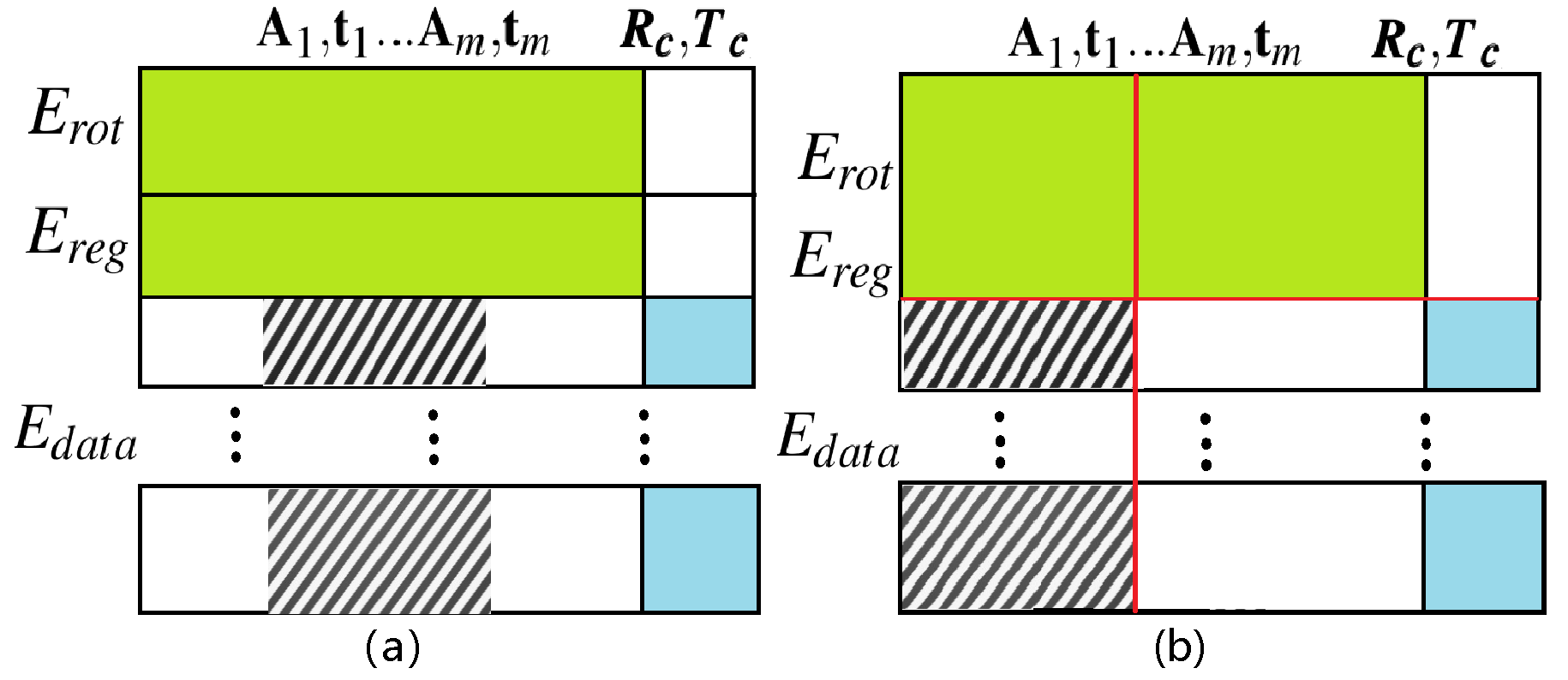}
		\caption{Left is a \textcolor{red}{Jacobian} matrix while the right is re-ordered \textcolor{red}{Jacobian}. Empty blocks are consisted of zero elements. } 
		\label{fig:Jacobian_fJacobian_fnew}
	\end{figure}

	It is natural to solve Eq. (\ref{TransformationMatrixFinalForm}) in batch. \textcolor{red}{As} the number of \textcolor{red}{vertices increases}, the dimension of Jacobian relating to state $\mathbf{\bm{\Phi}}$ increase dramatically. Fortunately, we can explore the fact that only partial ED nodes are connected to currently visible points, due to the limited field of view (FOV) of the camera. See Fig. \ref{fig:node_points_relation} as an example of constant FOV of depth image. The model keeps expanding while the target depth remains \textcolor{red}{in} small size. A typical ED node and target depth relationship is illustrated in Fig. \ref{fig:node_points_relation}(d); 2/3 of the nodes are not within depth FOV resulting no contribution to $E_{data}$. Fig. \ref{fig:Jacobian_fJacobian_fnew} shows a typical Jacobian of the cost function. In $E_{data}$ block, the shadow region indicates nodes connected to points (\textbf{PR} nodes) while zero block shows the nodes (\textbf{PI} nodes) connecting to inactive points. In this paper, we will exploit the sparity of PR nodes in the zero bocks.\par 
	The same sparsity also applies to Eq. (\ref{MatrixMC}). By rearranging matrix ${\bm{\Pi}}$ \textcolor{red}{from} Fig. \ref{fig:Jacobian_fJacobian_fnew}(a) to Fig. \ref{fig:Jacobian_fJacobian_fnew}(b), we achieve a new Jacobian with zero block. Using this new matrix, Eq. (\ref{TransformationMatrixFinalForm}) writes:

	\begin{equation}
	\label{matrixsplitting}
	\begin{aligned}
	E_{data}&=||{\mathrm{\mathbf{R}_c}}[{\bm{\Pi}}{\mathbf{\bm{\Phi}}}]^T+{\mathrm{{\mathbf{T}_c}}} \otimes \mathbf{1} -\mathbf{\tilde{P}}||_F^2\\
	&=||{\mathrm{\mathbf{R}_c}}[\left(
	\begin{array}{cc}
	{\bm{\Pi^{'}}}&\ {\mathbb{O}}\\
	\end{array}
	\right)
	\left(
	\begin{array}{c}
{\mathbf{\bm{\Phi}}}_1\\
	{\mathbf{\bm{\Phi}}_2}
	\end{array}
	\right)]^T+{\mathrm{{\mathbf{T}_c}}} \otimes \mathbf{1} 
	-\mathbf{\tilde{P}}||_F^2,\\
	\end{aligned}
	\end{equation}

	\noindent where ${\mathbf{\bm{\Phi}}}_1$ contains $\mathbf{A}_j$ and $\mathbf{t}_j$ of \textbf{PI} node set and ${\mathbf{\bm{\Phi}}_2}$ contains $\mathbf{A}_j$ and $\mathbf{t}_j$ of \textbf{PR} node set. $\bm{\Pi^{'}}$ is the subset of $\bm{\Pi}$ relating to \textbf{PI} nodes (the shadow region in Fig. \ref{fig:node_points_relation}).
	
	\vspace{0pt} 
	\hspace{0pt}
	\vspace{0pt} 
	\hspace{0pt}
	\begin{figure}[h]	
		\centering
		\subfloat{	
			\begin{minipage}[htpb]{0.22\textwidth}	
				\centering
				\includegraphics[width=1\linewidth]{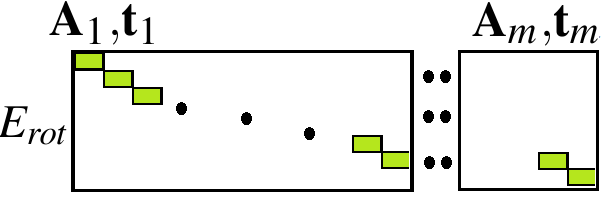}			
			\end{minipage}				
		}
		\subfloat{	
			\begin{minipage}[htpb]{0.22\textwidth}	
				\centering
				\includegraphics[width=1\linewidth]{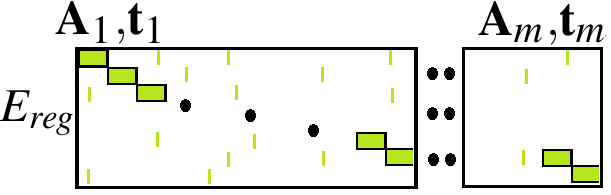}			
			\end{minipage}				
		}
		\caption{Left is Jacobian of $E_{rot}$ w.r.t all nodes. Right is Jacobian of $E_{reg}$ w.r.t all nodes.
		}
		\label{fig:Jacobian_Erot_Ereg}
	\end{figure}

%

	\subsection{Lossy Two level optimization}
	\label{two_level}
	\textcolor{red}{Explained} in Section \ref{session:Sparsity in ED node}, the size of the \textbf{PR} nodes is \textcolor{red}{almost} constant due to the limited size of depth image in expanding scenario. \textcolor{red}{For instance, the point cloud generated from  Hamlyn dataset \cite{giannarou2013probabilistic} (grabbed from monitor)} is $320\times240=76800$ at most. \textbf{As the model \textcolor{red}{grows}, the total number of nodes in ED graph is increasing but the number of nodes is almost constant}. Taking advantage of Eq. (\ref{matrixsplitting}), the optimization can be divided into two \textcolor{red}{levels}: the optimization of \textbf{PR} nodes ${\mathbf{\bm{\Phi}}}_1$ and the optimization of the rest \textbf{PI} nodes ${\mathbf{\bm{\Phi}}_2}$.\par 
	
	In specific, we first optimize (${\mathbf{\bm{\Phi}}}_1$, ${\mathrm{\mathbf{R}_c}}$ and ${\mathrm{{\mathbf{T}_c}}}$) by fixing ${\mathbf{\bm{\Phi}}_2}$ in (\textbf{Level I}) optimization, \textcolor{red}{to obtain an estimation of} ${\mathbf{\bm{\Phi}}}_1$, ${\mathrm{\mathbf{R}_c}}$ and ${\mathrm{{\mathbf{T}_c}}}$. \textcolor{red}{Then the value of the parameters obtained from \textbf{level I}, will be fixed in \textbf{Level II}, together with the two soft constraints $E_{rot}$ and $E_{reg}$ to optimize the parameter $\Phi_2$.} We explicitly enforce this idea by formulating following energy function:
	
	\begin{equation}
	\begin{split}
	\argmin\limits_{\mathbf{R}_c,\mathbf{T}_c,\mathbf{\bm{\Phi_1}}} &\ \omega_{rot}\tilde{E}_{rot}+\omega_{reg}\tilde{E}_{reg}+\omega_{data} E_{data}
	\end{split}
	\label{eq:energyfunction_1}
	\end{equation}
	
	\begin{equation}
	\begin{split}
	\argmin\limits_{\mathbf{\bm{\Phi}}_2} &\ \omega_{rot}E_{rot}+\omega_{reg} E_{reg}
	\end{split}
	\label{eq:energyfunction_2}
	\end{equation}
	
	Eq. (\ref{eq:energyfunction_1}) and Eq. (\ref{eq:energyfunction_2}) are the \textbf{Level I} and \textbf{Level II} energy functions, where $\tilde{E}_{rot}$ and $\tilde{E}_{reg}$ are \textcolor{red}{the curtailed} energy function of $E_{rot}$ and $E_{reg}$ containing $\mathrm{\bm{\Phi_1}}$. In other words, the size of Eq. (\ref{eq:energyfunction_1}) is only related to \textcolor{red}{the size of \textbf{PR} nodes}. \textcolor{red}{Therefore, the computational} complexity in \textbf{Level I} is \textcolor{red}{reduced} from $O(n^2)$ to constant $O(1)$ thanks to \textcolor{red}{the constant} size \textcolor{red}{of} ${\mathbf{\bm{\Phi}}}_1$. Although optimizing \textbf{Level II} is still $O(n^2)$, considering \textcolor{red}{the scale} of data term $E_{data}$ is far larger than the rest, \textcolor{red}{the computational cost} in \textbf{Level II} is relatively low. \textcolor{red}{Note} that the new strategy keeps the time consuming step \textbf{Level I} constant while \textbf{Level II} still $O(n^2)$. \textbf{But the size of \textbf{Level II} is almost negligible comparing with \textbf{Level I}}. \par
	
	\subsection{Connection with Marginalization and Information Loss}
	In this section, \textcolor{red}{we will draw the connection of the proposed two level optimization method with an exact marginalization based method. The analysis will show that the information loss is very low, illustrating the feasibility} of the decoupled optimization Eq. (\ref{eq:energyfunction_1}) and Eq. (\ref{eq:energyfunction_2}).\par 
	
	When generating Eq. (\ref{matrixsplitting}), the Jacobian \textcolor{red}{shown} in Fig. \ref{fig:Jacobian_fJacobian_fnew} is re-ordered by classifying $[\mathbf{A}_1, \mathbf{t}_1...\mathbf{A}_m, \mathbf{t}_m]$ into \textbf{PR} nodes ${\mathbf{\bm{\Phi}}}_1$ and \textbf{PI} nodes ${\mathbf{\bm{\Phi}}_2}$. The state in cost function $[\mathbf{R}_c,\mathbf{T}_c,\mathbf{A}_1,\mathbf{t}_1...\mathbf{A}_m,\mathbf{t}_m]$ are classified as $\mathbf{x}_c(\mathbf{R}_c,\mathbf{T}_c,\mathbf{A}_1,\mathbf{t}_1...\mathbf{A}_k,\mathbf{t}_k)$ and $\mathbf{x}_f(\mathbf{A}_{k+1},\mathbf{t}_{k+1}...\mathbf{A}_m,\mathbf{t}_m)$. Fig. \ref{fig:Jacobian_fJacobian_fnew} shows the Jacobian in the new order. The first two term are combined due to their sparsity because $E_{rot}$ and $E_{reg}$ are constraints between ED nodes, i.e. unrelated to feature points. The only full block in Fig. \ref{fig:Jacobian_fJacobian_fnew} is $E_{data}$ w.r.t $\mathrm{\bm{\Phi_1}}$ (shadow region), in specific $\frac{\partial \mathbf{J}_2}{\partial \mathbf{x}_c}$. \textcolor{red}{Let us write down the Jacobian and Hessian as,}\par 
	
	\begin{equation}
	\textcolor{red}{
	\boldsymbol{\mathcal{J}} = 
	\begin{bmatrix}
	\frac{\partial \mathbf{J}_1}{\partial \mathbf{x}_c}
	& 
	\frac{\partial \mathbf{J}_1}{\partial \mathbf{x}_f}
	\\
	\frac{\partial \mathbf{J}_2}{\partial \mathbf{x}_c}
	&
	\mathbb{O}	 
	\end{bmatrix}
	\stackrel{\mathrm{def}}{=}
	\begin{bmatrix}
	\mathbf{J}_{1c}	&	\mathbf{J}_{1f}
	\\
	\mathbf{J}_{2c}	&	\mathbb{O}	 
	\end{bmatrix}	 
}
	\end{equation}
	
\begin{equation}
\textcolor{red}{
\boldsymbol{\mathcal{H}} 
= 
\begin{bmatrix}
\mathbf{J}_{1c}^T \mathbf{J}_{1c} + \mathbf{J}_{2c}^T \mathbf{J}_{2c}	&	\mathbf{J}_{1c}^T \mathbf{J}_{1f}
\\
\mathbf{J}_{1f}^T \mathbf{J}_{1c}	&	\mathbf{J}_{1f}^T \mathbf{J}_{1f}	 
\end{bmatrix}
\stackrel{\mathrm{def}}{=}
\begin{bmatrix}
\mathbf{\Lambda}_{cc}	&	\mathbf{\Lambda}_{cf}
\\
\mathbf{\Lambda}_{cf}^T	&	\mathbf{\Lambda}_{ff}	 
\end{bmatrix}	 
}
\end{equation}

	The density of $\frac{ \partial \mathbf{J}_2 }{  \partial \mathbf{x}_c}$ is the only dense block in Hessian $\boldsymbol{\mathcal{H}}$. Let us use Schur complement \cite{thrun2005probabilistic} from classic rigid SLAM and separate the optimization as follows:

\begin{equation}
\textcolor{red}{
\label{Eq_Hessian_1}
\begin{bmatrix}
\mathbf{\Lambda}_{cc}	&	\mathbf{\Lambda}_{cf}
\\
\mathbf{\Lambda}_{cf}^T	&	\mathbf{\Lambda}_{ff}	 
\end{bmatrix}	 
\begin{bmatrix}
\mathbf{x}_c  \\  \mathbf{x}_f
\end{bmatrix} 
=
\begin{bmatrix}
\mathbf{J}_{1c}^T	\mathbf{F} +	\mathbf{J}_{2c}^T \mathbf{F}
\\
\mathbf{J}_{1f}^T	\mathbf{F}	 
\end{bmatrix}	 
\stackrel{\mathrm{def}}{=}
\begin{bmatrix}
\mathbf{y}_c  \\  \mathbf{y}_f
\end{bmatrix} }
\end{equation}
\begin{equation}
\textcolor{red}{
\label{Eq_Hessian_2}
\begin{bmatrix}
\mathbf{\Lambda}_{cc} - \mathbf{\Lambda}_{cf}
\mathbf{\Lambda}_{ff}^{-1} \mathbf{\Lambda}_{cf}^T
&	\mathbb{O}
\\
\mathbf{\Lambda}_{cf}^T	&	\mathbf{\Lambda}_{ff}	 
\end{bmatrix}	 
\begin{bmatrix}
\mathbf{x}_c  \\  \mathbf{x}_f
\end{bmatrix} 
=
\begin{bmatrix}
\mathbf{y}_c - \mathbf{\Lambda}_{cf}
\mathbf{\Lambda}_{ff}^{-1} \mathbf{y}_f
\\  \mathbf{y}_f
\end{bmatrix} 
}
\end{equation}
	
\vspace{0pt} 
\hspace{0pt}
\begin{figure}
	\centering 
	\includegraphics[width=0.3\textwidth]{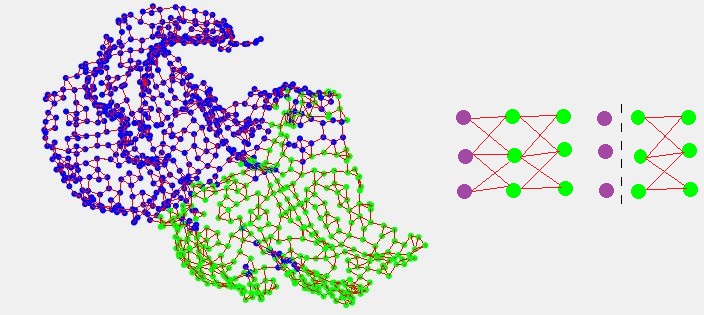}
	\caption{Left shows two types of nodes and edges. Green nodes are the \textbf{PR} nodes and purple nodes are \textbf{PI} nodes. The right indicates the connections of \textbf{PR} and \textbf{PI} nodes. Left is the full connection while the \textbf{PR} and \textbf{PI} connections are cut in our \textbf{Level I} optimization} 
	\label{fig:2types_of_nodes_visualize}
\end{figure}
\vspace{0pt} 
\hspace{0pt}
\begin{figure}
	\centering 
	\includegraphics[width=0.38\textwidth]{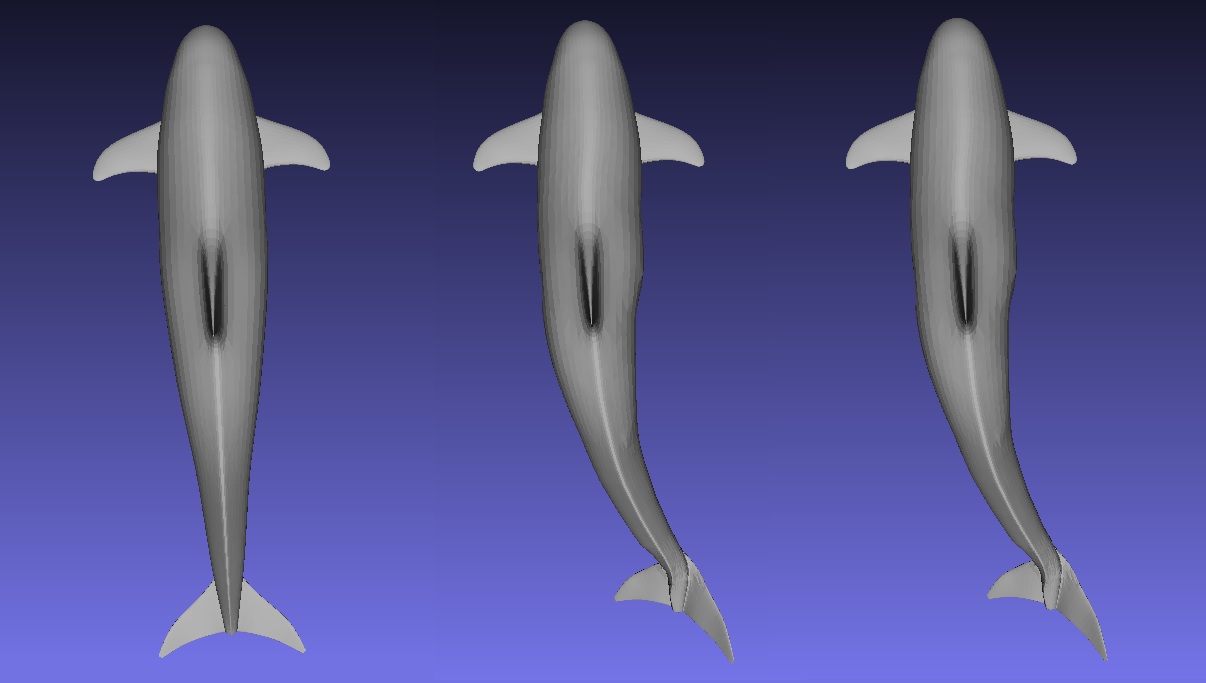}
	\caption{Qualitative comparisons of our strategy and original ED based deformation. The first shape is the original dolphin mesh. We show the result of deformed shape (the last) along with the result of classical ED deformation (middle).} 
	\label{fig:ED_deformation_comparison}
\end{figure}

\begin{figure}
	\centering 
	\includegraphics[width=0.43\textwidth]{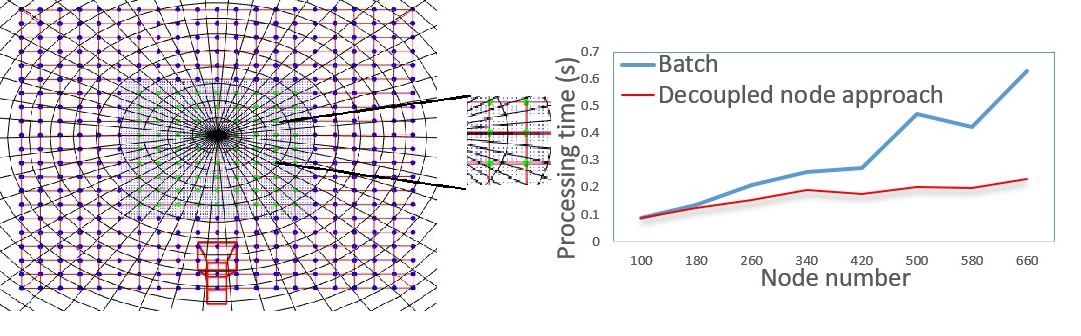}
	\caption{An example model with static camera. One step optimization step from plain surface model (tiny dots) to warped surface (grid). \textbf{PI} nodes are in blue and \textbf{PR} nodes are in green. } 
	\label{fig:simulation_experiment}
\end{figure}
By using Schur complement, we can solve $\mathbf{x}_c$ independent of $\mathbf{x}_f$. The computation of $\mathbf{x}_c$ (including $\mathrm{\mathbf{R}_c}$, $\mathrm{\mathbf{T}_c}$ and ${\mathbf{\bm{\Phi}}}_1$) is constant (explained in Section \ref{two_level}). After solving $\mathbf{x}_c$, the optimization of $\mathbf{x}_f$ is quite cheap as the $\bm{\Lambda}_{ff}=(\frac{\mathbf{J}_1}{\partial \mathbf{x}_f})^T(\frac{\mathbf{J}_1}{\partial \mathbf{x}_f})$ is only related to $E_{rot}$ and $E_{reg}$. The sparsity of Hessian and small number of nodes ($n \gg m$) makes the time of solving $\mathbf{x}_f$ much less.\par

	Fig. \ref{fig:Jacobian_Erot_Ereg} is the Jacobian of $E_{rot}$ and $E_{reg}$ relating to all nodes. The first term $E_{rot}$ is the sum of error of affine transformation (Eq. (\ref{Eq_E_rot})) making the Jacobian strictly diagonal. The second term $E_{reg}$ defines the transformation error among ED nodes (Eq. (\ref{RegulationConstraint})). The major part $\mathbf{A}_j(\mathbf{g}_k-\mathbf{g}_j)+\mathbf{g}_j+\mathbf{t}_j$ is also within one node $j$ except the very last $-(\mathbf{g}_k+\mathbf{t}_k)$. The last variable $\mathbf{t}_k$ makes the Jacobian not strictly diagonal (please refer to Fig. \ref{fig:Jacobian_Erot_Ereg}(b)). We come up with an idea that by ignoring Jacobian of $\mathbf{t}_k$, and reordering two diagonal Jacobians, $[\frac{\partial {\mathcal{J}}_1}{\partial \mathbf{x}_c}
	\frac{\partial {\mathcal{J}}_1}{\partial \mathbf{x}_f}]$ becomes diagonal and $\boldsymbol{\Lambda}_{cf} = (\frac{\partial {\mathcal{J}}_1}{\partial \mathbf{x}_c})^T(\frac{\partial {\mathcal{J}}_1}{\partial \mathbf{x}_f})=0$. \textcolor{red}{In this case, Eq. (20) can be separated into two equations, with respect to $\mathbf{x}_c$ and $\mathbf{x}_f$ respectively, which also correspond to the Level I and Level II of the proposed method. In a word, our two level lossy method can be considered as an approximation of Eq. (21), under the assumption that $\boldsymbol{\Lambda}_{cf} =0$ roughly stands.}

	Fig. \ref{fig:2types_of_nodes_visualize} visualizes the feasibility of the lossy decoupled optimization approach in geometry. \textcolor{red}{\textbf{PR} nodes (green)} are the only nodes connected to visible points and contribute to $E_{data}$. \textcolor{red}{All \textbf{PI} nodes (purple)} merely share edges with \textcolor{red}{\textbf{PR} nodes} and are passively deformed according to \textcolor{red}{the} behaviors of \textbf{PR} nodes. Equivalently, the inter-nodes relations in the Jacobian of $E_{reg}$ (Fig. \ref{fig:Jacobian_Erot_Ereg}(b)) shows these connections (Fig. \ref{fig:Jacobian_Erot_Ereg}(b)). In view of this, our lossy decoupled optimization approach first optimize \textcolor{red}{\textbf{PR}} nodes and then estimate \textcolor{red}{\textbf{PI}} nodes.\par 
	
	\textbf{In conclusion, solving energy function Eq. (\ref{energyfunction}) by ignoring Jacobian of $\mathbf{t}_k$ is equivalent to the proposed decoupled optimization in Eq. (\ref{eq:energyfunction_1}) and Eq. (\ref{eq:energyfunction_2}).}\par

The information loss of the proposed approach is relatively low. Fig. \ref{fig:2types_of_nodes_visualize} illustrates the connection between \textbf{PR} nodes and \textbf{PI} nodes is weak on the boundary, in contrast with the dense connections among \textbf{PR} nodes. It also demonstrates how the connection between \textbf{PR} and \textbf{PI} nodes are removed in \textbf{Level I}, and the connection between $\frac{\partial \mathbf{J}_1}{\partial \mathbf{x}_c}$ and
$\frac{\partial \mathbf{J}_1}{\partial \mathbf{x}_f}$ are removed resulting in $\bm{\Lambda}_{cf}=(\frac{\partial \mathbf{J}_1}{\partial \mathbf{x}_c})^T(\frac{\partial \mathbf{J}_1}{\partial \mathbf{x}_f})=\mathbb{O}$. \textcolor{red}{The information} between two \textbf{PR} nodes is strong while \textcolor{red}{that among the} \textbf{PI} nodes is weak. The weak information is neglected in \textbf{Level I}, attributing to relatively low information loss in optimization process.\par

	\section{Results and discussion}
	Our goal is \textcolor{red}{to} have both qualitative assessments and quantitative comparisons between the original ED \textcolor{red}{graph optimization and the proposed method}. For qualitative comparison, we \textcolor{red}{show} the sacrificed accuracy \textcolor{red}{has} few \textcolor{red}{impacts} on final \textcolor{red}{reconstructed map}. A dolphin model is downloaded from turbosquid (https://www.turbosquid.com). w.r.t quantitative test in SLAM, \textcolor{red}{both methods are compared} on \textcolor{red}{a synthetic dataset and datasets from} \cite{giannarou2013probabilistic}, where we choose three in-vivo stereo videos with deformation and rigid scope movement. \textcolor{red}{The experiments are conducted on the same hardware and sorfware setting of MIS-SLAM \cite{song2018mis}}. The module of state estimation in MIS-SLAM is modified to \textcolor{red}{the proposed approach}. \textcolor{red}{Note that an iterative solver, i.e. the preconditioned conjugate gradient method, is employed to solve the resulting linear systems, as it provides a way of parallel computing on GPU.} \par 
	

	\subsection{Qualitative ED deformation comparisons}
	\label{Quanlitative_comparison}
	
	Fig. \ref{fig:ED_deformation_comparison} shows a qualitative comparison. It is not aiming \textcolor{red}{as a proof of the superiority of the proposed method over the original ED graph method}. \cite{sumner2007embedded} has already claimed real-time implementation on CPU as well as very nice results. Aiming at speeding up deformable SLAM application, the qualitative result of our lossy \textcolor{red}{decoupled} approach is not comparable to the batch estimation of ED. However, we want to illustrate that the proposed method can achieve a similar result and the difference is not visible to the naked eye or difficult to make out. Fig. \ref{fig:ED_deformation_comparison} confirms that the deformed shape performed by our approach do actually have \textcolor{red}{similar result}. \textcolor{red}{It looks similar to the original ED graph partially due to simple topology of dolphin.} Other complex models like human results \textcolor{red}{in} a visible but not very apparent difference. \textcolor{red}{Although the \textcolor{red}{decoupled} optimization works well, the results can be much worse than ED when the nodes are too sparse. The deformation is dependent on \textbf{PR} nodes and \textcolor{red}{the insufficiency} of \textbf{PR} nodes (or nodes in conventional one step ED) causes vertex deformation \textcolor{red}{that the expected target is not reached.}}\par

\begin{figure}
	\centering 
	\includegraphics[width=0.48\textwidth]{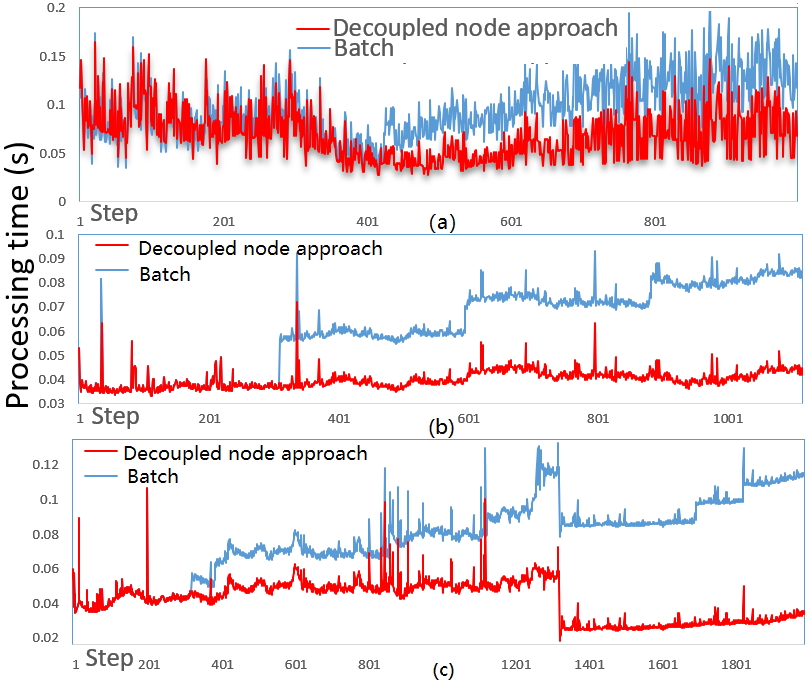}
	\caption{Processing time comparsions of model 6 (a), 20 (b) and 21 (c) in Hamlyn dataset. Blue lines are the batch optimization and red lines are our nodes decoupled optimization. We cannot illustrate \textbf{Level I} and \textbf{Level II} separately due to time consumption of \textbf{Level II} is extremely low.} 
	\label{fig:timecomplexitycomparisons}
\end{figure}
\begin{figure}
	\centering 
	\includegraphics[width=0.48\textwidth]{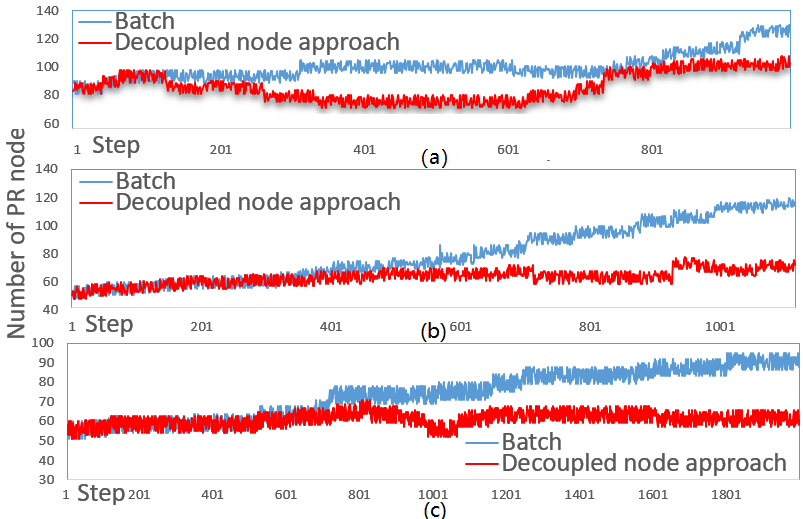}
	\caption{Optimizing nodes comparisons in first level computation of model 6 (a), 20 (b) and 21 (c) in Hamlyn dataset. The red lines are the result of our decoupled optimization strategy while the blue lines are the original batch strategy. } 
	\label{fig:2types_of_nodes}
\end{figure}
	
	\subsection{Time complexity comparisons}	
	\textcolor{red}{Fig. \ref{fig:simulation_experiment} show a tiny one-step simulation and the result. Tractable time consumption remains small because \textbf{PR} node does not change.}\par 
	\textcolor{red}{We compare the original MIS-SLAM \cite{song2018mis} with the improved version}. Fig. \ref{fig:timecomplexitycomparisons} illustrates the running time for three Hamlyn datasets (model 6, 20 and 21). In all scenarios \textcolor{red}{decoupled} optimization \textcolor{red}{yields} better efficiency than batch processing especially in case of long term process (the last dataset in Fig. \ref{fig:timecomplexitycomparisons}). In the first few steps, the robot is steady and ED graph is not expanding significantly. This attributes to the similar processing time in the first few hundred steps. When the robot moves, the ED graph expands intensively and processing time increases abruptly in state optimization. In view of this, by limiting the node graph, decoupled optimization keeps time consumption stable due to the constant \textbf{PR} node scale.    The attached video shows the range of movement in model 6 is much smaller than model 20 and 21. That's the main reason the proposed method does not improve model 6 too much. As the environment gets larger, our approach keeps much lower time consumption.\par

	\subsection{Accuracy comparisons}	
	
	The lossy \textcolor{red}{decoupled} optimization strategy inevitably attributes \textcolor{red}{the loss of} accuracy. Section \ref{Quanlitative_comparison} shows the quality of the deformed map is well preserved in ED deformation process. Moreover, we compare the proposed method and the original one on the same parameters and weights of terms in SLAM application (MIS-SLAM). Different from arbitrary key points matching in ED deformation formulation (Eq. (\ref{E_dataterm})), in SLAM application the $E_{data}$ is in the form of model-to-depth scan matching like point-to-plane ICP. For quantitative validation, we measure the point-to-point distance of deforming map and target scan. \textcolor{red}{For direct validation to ground truth, three synthetic data (heart, right kidney, and stomach) are generated by deliberately deforming models from CT scanned phantom. As a compliment, the three laparoscopy datasets from Hamlyn are tested, but only the back-projection error in each iteration is available since there's no ground truth.} In the batch approach, the average distance of back-projection registration of the three simulation scenarios is 0.18mm (model 6), 0.13mm (model 20) and 0.12mm (model 21). While dataset with ground truth (Hamlyn dataset 10/11) achieves 0.08mm, 0.21mm (Average errors). With our decoupled optimization approach, we achieve 0.31mm (model 6), 0.26mm (model 20), 0.22mm (model 21) and 0.14mm, 0.29mm errors. The attached video shows no big difference in terms of structure and texture. \par
	
	We also test the average error. On top of the in-vivo dataset, synthetic data are generated. Please refer to the attached video for more details. Roughly speaking, the proposed method sacrifices $32\%$ accuracy in exchange for $50\%$ speed gains.\par

	\section{Conclusion}
	We propose a decoupled approach for ED graph optimization that reduces computational complexity from $O(n^2)$ to near $O(1)$. The decoupled optimization structure achieves faster computation in scenario of expanding environment. Our strategy sacrifices small amount of accuracy in exchange for near-constant processing speed. \textcolor{red}{The constant computation complexity of the lossy strategy should have great potential in ED graph based SLAM in unbounded map scenario.}\par 
	\textcolor{red}{The node marginalization strategy in this paper, however, is straightforward, in the sense that it only classifies nodes based on the node-vertex connectivity. It's reasonable because different from the pose graph, ED graph is parallelized in GPU since the time consumption requirement is more strict. However, it remains to be of great interest to test if pose graph pruning method like Kullback–Liebler divergence outperforms the proposed work while remains acceptable consumption in GPU environment.}\par

	




	\bibliographystyle{ieeetr}
	\bibliography{reference}   

\end{document}